\documentclass[a4paper,fleqn]{cas-dc}

\usepackage[numbers]{natbib}

\def\tsc#1{\csdef{#1}{\textsc{\lowercase{#1}}\xspace}}
\tsc{WGM}
\tsc{QE}

\begin{document}
\let\WriteBookmarks\relax
\def\floatpagepagefraction{1}
\def\textpagefraction{.001}


\shortauthors{Lim and Lee}

\title [mode = title]{Intelligent Repetition Counting for Unseen Exercises: A Few-Shot Learning Approach with Sensor Signals}

\tnotemark[1]

\tnotetext[1]{This work is the result of the research project funded by Samsung Electronics Co., Ltd., and was also supported by the National Research Foundation of Korea (NRF) under Grant 2021R1F1A1061093.}

\author[1]{Yooseok Lim}[orcid=0009-0005-5962-5085]

\ead{seook6853@soongsil.ac.kr}

\credit{Writing - original draft, Methodology, Software, Data curation}

\affiliation[1]{organization={Department of Industrial and Information Systems Engineering, Soongsil University},
            addressline={369, Sangdo-ro, Dongjak-gu}, 
            city={Seoul},
            postcode={06978}, 
            country={Republic of Korea}}

\author[2]{Sujee Lee}[orcid=0000-0003-1060-2067]

\cormark[1]

\ead{sujeelee@skku.edu}

\credit{Writing - review \& editing, Project administration, Conceptualization, Methodology}

\affiliation[2]{organization={Department of Systems Management Engineering, Sungkyunkwan University},
            addressline={2066, Seobu-ro, Jangan-gu}, 
            city={Suwon},
            postcode={16419}, 
            country={Republic of Korea}}

\cortext[1]{Corresponding author}



\begin{abstract}
Sensing technology has significantly advanced in automating systems that reflect human movement, particularly in robotics and healthcare, where it is used to automatically detect target movements. This study develops a method to automatically count exercise repetitions by analyzing IMU signals, with a focus on a universal exercise repetition counting task that counts all types of exercise movements, including novel exercises not seen during training, using a single model. Since peak patterns can vary significantly between different exercises as well as between individuals performing the same exercise, the model needs to learn a complex embedding space of sensor data to generalize effectively.

To address this challenge, we propose a repetition counting technique utilizing a deep metric-based few-shot learning approach, designed to handle both existing and novel exercises. By redefining the counting task as a few-shot classification problem, the method is capable of detecting peak repetition patterns in exercises not seen during training. The approach employs a Siamese network with triplet loss, optimizing the embedding space to distinguish between peak and non-peak frames. 
Evaluation results demonstrate the effectiveness of the proposed approach, showing an 86.8\% probability of accurately counting ten or more repetitions within a single set across 28 different exercises. This performance highlights the model's ability to generalize across various exercise types, including those not present in the training data. Such robustness and adaptability make the system a strong candidate for real-time implementation in fitness and healthcare applications
\end{abstract}


\maketitle

\section{Introduction}\label{sec:introduction}

As the adoption of smart devices and the advancements in sensor technology embedded in mobile devices continue to grow, vast amounts of sensor data are being collected and utilized across various fields. This sensor data is employed for pattern analysis and information extraction in diverse areas such as medical care, physical education, and daily life. For instance, continuous monitoring systems are effective for improving the health of patients with Parkinson's disease \cite{parkinson} and cardiovascular conditions \cite{har_rc_rehabilitation}, which often require significant attention from physicians. Wearable sensors can simplify the process by automatically estimating and tracking patients' health states, thereby reducing human effort.

In recent years, sports healthcare systems have leveraged IMU sensors to offer a range of services \cite{har_rc_rehabilitation, har_rc_workout, har_arm, har_continuous, har_locomotion}. Wearable devices such as fitness trackers and smartwatches facilitate automatic exercise tracking and provide coaching functions. However, most existing features are limited to tracking exercise duration or measuring speed and heart rate. Despite technological advancements, commercial wearable devices still lack the capacity to record all critical information for exercise tracking. 

Numerous studies have aimed to develop more accurate activity recognition or exercise counting models. Especially in recent works, deep learning studies have utilized IMU sensor data across various domains. From a healthcare industry perspective, models have been proposed for limb movement rehabilitation \cite{har_limb}, and fall risk detection \cite{har_fall_risk}. In athletics-related services, models have been developed for free weight exercise monitoring \cite{har_free_weight}, gym activity monitoring \cite{har_gym}, lunge performance evaluation \cite{har_lunge}, exercise feedback \cite{har_rc_exercise}, and intensity recognition in strength training \cite{har_strength}.

However, most previous studies aim to classify the target activities, whereas the repetition counting task is more challenging due to the short and variable duration of unit activities. Moreover, their practical applications are hindered by models that only cover specific exercises, which are included in their dataset, and necessitate attaching multiple sensors to various body parts, causing inconvenience for users. Given the countless exercises in weight training and home workouts, as well as the potential for new workouts to emerge, it is nearly impossible to collect comprehensive training data that encompasses all exercise types. Therefore, this study aims to develop a universal repetition counting model using a few-shot learning algorithm, capable of accurately counting repetitions across a wide range of weight training and home fitness exercises, including exercises not present in the training dataset.

To address this challenge, we employed a metric-based algorithm that learns a universal embedding, which captures the common characteristics across various types of exercises. This embedding enables the model to generalize beyond the specific movements seen during training and recognize novel movements outside the training dataset. In our previous study \cite{har_sensor_fusion}, deep learning models were developed for repetition counting of 30 different types of exercises, utilizing an extensive exercise dataset privately collected through a series of experiments with a single head-mounted IMU sensor. Building upon this previous work, the current study aims to create a more generalized model for counting repetitions of new exercises using a few-shot learning technique with a Siamese network and triplet loss function \cite{facenet}.

The primary contributions of this study are as follows: firstly, it is the first to apply a metric-based learning approach to exercise repetition counting. Secondly, it introduces a real-time capable system. Lastly, the effectiveness of our method has been validated across 28 different exercises.

The remainder of this paper is organized as follows: Section \ref{sec:2} reviews related work on repetition counting using IMUs and few-shot learning. In Section \ref{sec:3}, the problem description, repetition counting framework, and various methods are explained. Section \ref{sec:4} covers the details of the dataset and the experimental setup. The prediction results are examined in Section \ref{sec:5}. Section \ref{sec:6} offers a discussion on the implications of the obtained results, addresses the limitations of the study, and suggests plans for future research. Finally, the conclusion is drawn in Section \ref{sec:7}.

\section{Related work }\label{sec:2}

\subsection{Repetition counting}
An early study on smartphone-based recognition systems \cite{har_arm} conducted research closely related to repetition counting. In the study, a smartphone was attached to the upper arm to collect sensor data, with the system divided into three processes: segmentation, recognition, and counting. The researchers determined heuristic thresholds and appropriate segmentation sizes, performed activity recognition using machine learning, and applied a rule-based statistical peak detection algorithm. Subject-independent training achieved 85\% segmentation accuracy and 94\% recognition accuracy across ten classes. Reco-Fit \cite{recofit} improved robustness and accuracy compared to \cite{har_arm} by employing a smartwatch sensor and implementing learned segmentation for enhanced robustness and scalability. The study introduced dimensionality reduction for orientation-invariant analysis and improved counting accuracy by incorporating false peak rejection and a more comprehensive repetition counting approach.

Similarly, another study \cite{har_rc_workout} mounted a sensor on the chest to analyze four activities (push-ups, sit-ups, squats, and jumping jacks), achieving an average detection accuracy of 97.9\% across the four workout types. Soro \textit{et al}. \cite{har_rc_exercise} examined ten weight training activities using IMUs, attaching the sensor to the wrist and using a 1D-convolutional neural network for counting. The error distribution of repetition counting was 73.5\%, 17.3\%, 2.2\%, and 7.0\%, depending on the number of errors ($0,1,2,>2$). Prabhu \textit{et al}. \cite{har_rc_rehabilitation} categorized ten activities into upper-body and lower-body movements, analyzing the data using statistical and neural network methods. The statistical technique showed high performance for upper-body motion but low performance for lower-body motion, whereas the neural network displayed high performance for both categories.

However, these previous approaches had limitations in terms of complexity and scalability in counting. They required three steps for counting, with errors in each step affecting subsequent steps and the complexity of the pipeline, resulting in decreased accuracy. In terms of scalability, statistical methods for peak detection define a peak or local maximum as any sample whose two direct neighbors have a smaller amplitude. The neighbor's distance is a crucial hyperparameter that should be heuristically designated for each exercise. However, analyzing all muscle exercises is time-consuming, and applying these methods to sports with no clear local maximum is challenging.

In contrast to these multi-step approaches, the counting model proposed in this study utilizes deep learning to learn peak features, eliminating the need for heuristic adjustments. It compresses the counting pipeline into a single step rather than two or three steps and conducts extensive experiments with a dataset of 28 sports events. The model is unique in its use of a single sensor placed above the ear, as opposed to the chest and wrist positions typically chosen in previous studies. Since head movement varies more significantly among individuals than other joint movements, subject variation is relatively larger. Additionally, using a single sensor creates a more sensitive environment, as the system is more susceptible to variations in sensor readings due to individual differences, noise, or other factors. Employing multiple sensors could potentially provide more information and help the system better handle these variations, but it may also increase complexity and inconvenience for the user. Unlike previous studies that focused on smartwatches as target devices for models, this study confirms that data-based sensor functions can be applied to embedded systems in other types of smart devices, such as hairbands and earphones.

Notably, the proposed counting model operates robustly for all workouts, not limited to specific ones. Existing studies have independently constructed models for exercises and evaluated performance through individual models, leading to limitations in terms of generalization. In contrast, this model counts the repetitions of all exercises using a single model and measures the repetitions regardless of the event, resulting in strong generalization capabilities. This framework offers a practical solution for addressing repetition counting, as it can be applied to a wide range of exercises without being constrained to specific ones.

\subsection{Few-shot learning and metric-based learning}
Few-shot learning, a subfield of machine learning, focuses on creating models capable of generalizing from limited training data. The primary focus is on classification, specifically predicting class labels for new instances using a small number of labeled examples per class. Typically, this involves an N-way M-shot task, where N represents the number of classes and M represents the number of labeled examples per class. The goal is to train a model to accurately classify new instances using only M examples per class, which significantly reduces the dependency on large labeled datasets for training.

Metric-based learning is one of the techniques used in few-shot learning, where data is learned and classified based on distance. The core objective is to embed data in a high-dimensional space so that similar data points are positioned close to each other, while different data points are positioned farther apart. 
FaceNet \cite{facenet} is a representative approach for solving few-shot tasks, particularly in the context of face recognition. It handles binary classification and employs triplet loss to learn an embedding space, ensuring that the distance between feature vectors of the same class (positive pair) is close, while those of different classes (negative pair) are far apart.
Matching network \cite{matching_network} obtains the similarity of the support set and query set through attention mechanisms and take the class with the highest similarity as the result. Matching network utilizes the episodic training method to effectively use a small multi-class training set.

There are several studies that deal with human exercise recognition based on few-shot learning \cite{rc_few_shot,har_few_shot_fu1,har_few_shot_fu2}. For example, Fu \textit{et al}. \cite{har_few_shot_fu1,har_few_shot_fu2} utilized various few-shot learning methods to classify eight exercise movements. By employing Siamese networks \cite{facenet}, ProtoNet \cite{proto}, and LocalNet \cite{local_net}, they compared the similarity of exercise movements and also propose a method to reduce the distribution difference between lab and real environments using domain adaptation techniques. Nishino \textit{et al}. \cite{rc_few_shot} used a knowledge transfer learning-based few-shot approach for exercise repetition counting, demonstrating that the fine-tuning process with a small amount of user data can significantly improve performance. Additionally, it incorporates the auto-correlation technique to determine the start and end points of windows. According to our investigation, this is the only study that addresses the same problem as ours, specifically focusing on human exercise repetition counting using a few-shot approach. 

However, there are significant differences between their study and ours. Their approach, which uses the entire repetition signal as input, is unsuitable for real-time applications due to the large computational load. Furthermore, their use of transfer learning, while effective in certain cases, struggles to capture the extensive variability within the embedding space. In contrast, our model processes smaller segments of data using a sliding window method, making it more efficient for real-time use. Additionally, our metric-based learning approach better handles intra-class variations by optimizing the embedding space through positive and negative pairs, offering more flexibility and generalization.

This makes the triplet loss-based Siamese network, a metric-based approach particularly well-suited for the task of classifying binary peak and non-peak classes. Its pair-based training algorithm provides flexible adaptation to new movements and high generalization capability in the embedding space, enabling robust performance across various exercises.

\section{Materials and methods}\label{sec:3}
This section illustrates the repetition counting system for a few-shot setting and the key methodologies. Specifically, it covers the system configuration and the overall framework. Additionally, it explains the Siamese network and transition counting, which are the primary methodologies of this study.

\begin{figure}[t]
    \centering
    \includegraphics[width=1.0\linewidth]{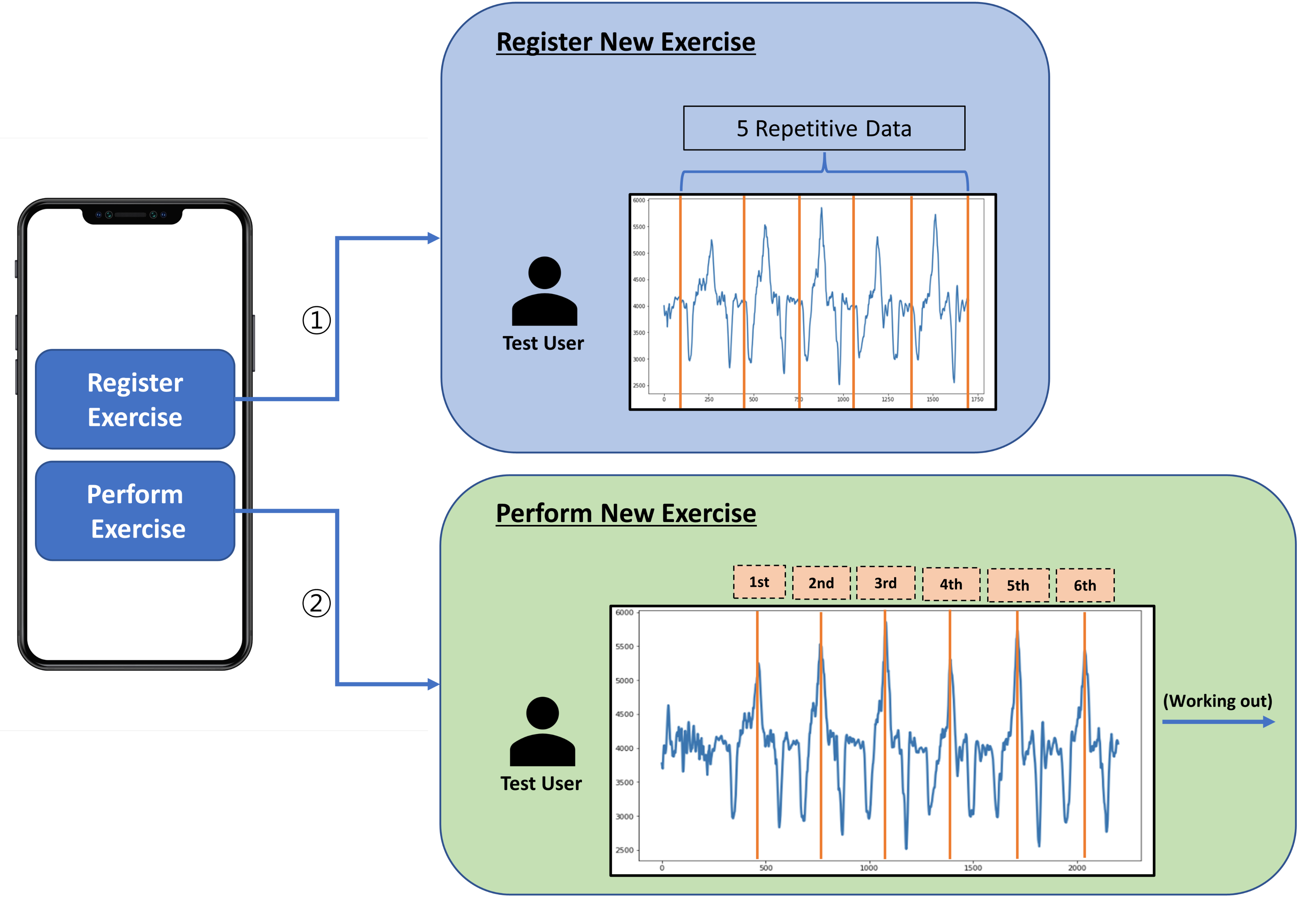}
    \caption{Overview of the proposed exercise repetition counting system}
    \label{fig: pd}
\end{figure}

\subsection{Proposed system description}
\label{sec:3-a}

The primary purpose of the proposed model is to count the number of repetitions of exercises when a user of the service performs strength training or home fitness. Furthermore, the study focuses on counting repetitions of new exercises that are not included in the training data.

The proposed model can be used in an application for exercise repetition counting with the following process. When a user performs a set of exercises, sensor data capturing head movement is collected using an IMU sensor attached above either side of the ears to recognize a user's movement. When the type of the user's exercise is one of the typical exercises included in the model training step (e.g., squat, deadlift), the developed model analyzes the sensor patterns in real time to detect the peak points and counts the repetitions. If the user wants to conduct a new exercise, the registration process, which collects five repetitive patterns of the exercise, is required before performing the new exercise (see Fig. \ref{fig: pd}). Users perform the peak movement of the exercise according to the timing guided by the system, which is utilized to perceive a peak signal. It is a significant procedure to create pair sets of new exercises. After the new exercise registration is completed, the repetitions are counted using the model developed through the few-shot learning.

\begin{figure*}[t]
    \centering
    \includegraphics[width=1.0\linewidth]{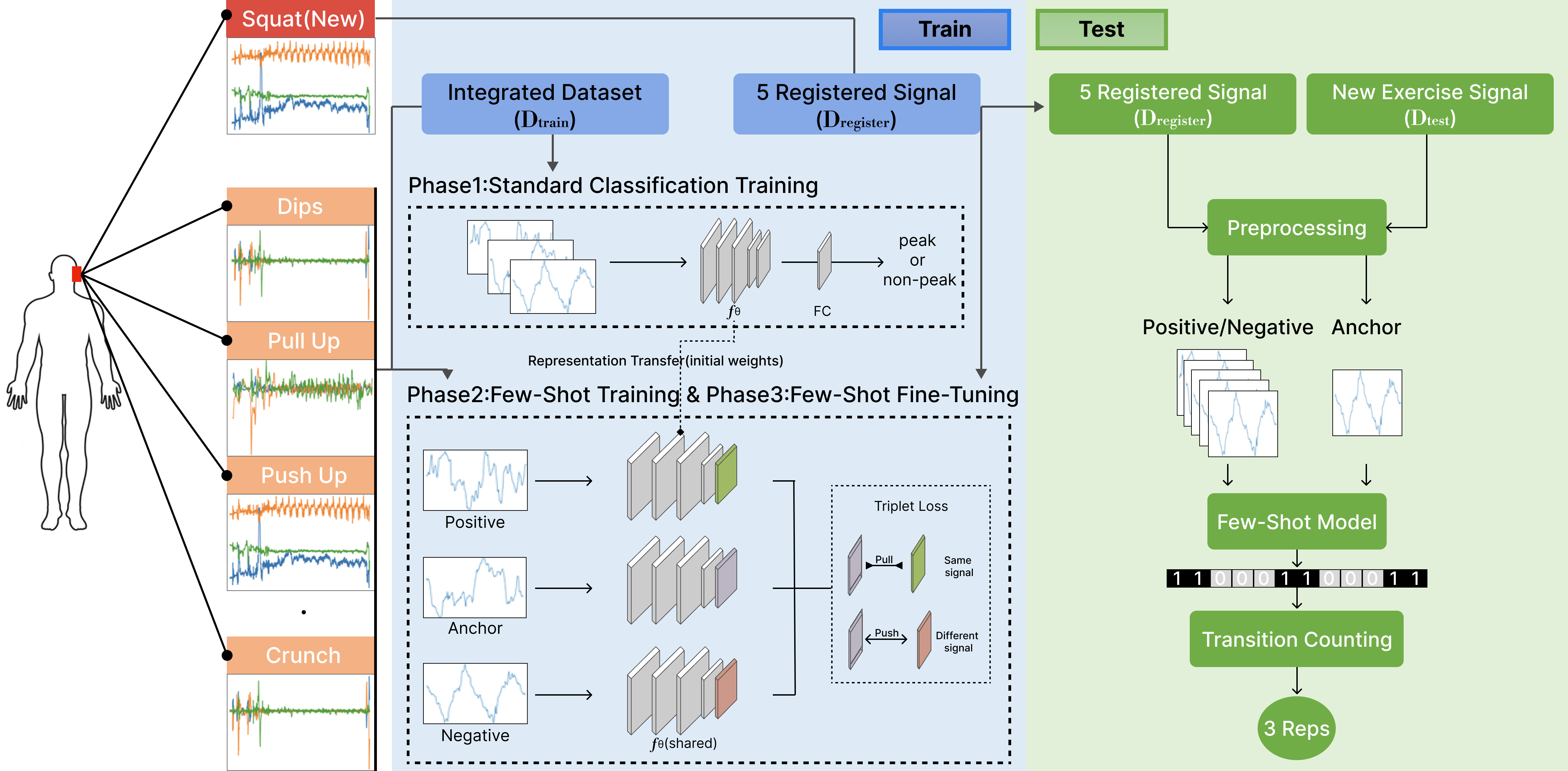}
    \caption{Overview of Few-shot Learning-based Modeling Process for Repetition Counting}
    \label{fig: overview}
\end{figure*}

\subsection{Siamese network, Triplet pair and loss}
\label{sec:3-c}
We use a Siamese network, specifically composed of two networks that share the same weights. It takes two inputs, and each input is transformed into an embedding vector through a network with identical weights. The network measures the similarity between two embedding vectors and learns the relative distance between samples.

In this study, triplet pairs are used as input to the Siamese network. These pairs help the model learn the distance between different classes in the embedding space. The anchor ($x^a$) is the reference sample representing the target signal, the positive ($x^p$) is a sample from the same class, and the negative ($x^n$) is from a different class.

Triplet loss helps the model minimize the distance between anchor-positive pairs while maximizing the distance between anchor-negative pairs. This creates an embedding space where similar samples are close together, and dissimilar samples are further apart.

\subsection{Transition counting}
\label{sec:3-d}
Transition counting serves as a post-processing step, converting the classification results into discrete values to determine the number of repetitions. After the continuous IMU sensor data has been segmented using a sliding window approach, and each window has been classified by the model as either peak (1) or non-peak (0), the transition counting function is applied. This post-processing function works by identifying transitions between the classified classes. Specifically, when the model outputs a sequence of 1s and 0s, where 1 indicates the peak of a repetition, the transition counting function detects a sequence of consecutive 1s, treating them as one complete repetition. By accumulating these sequences, the total number of repetitions is determined, ensuring that the model’s predictions are effectively transformed into meaningful repetition counts for practical use.

\subsection{Few-shot repetition counting framework}
\label{sec:3-e}

Fig. \ref{fig: overview} provides a high-level overview of the process. Our system is divided into training and testing procedures.

In the training step, the goal is to learn a generalized embedding model ($f_\theta$) that can classify peak and non-peak points for new exercises. First, we perform standard classification training (Phase 1), where a binary classification network is trained on all base classes. After training, the final fully-connected layer is removed, and the remaining weights form the initial model $f_\theta$, providing an embedding space with strong transferability for few-shot learning. 
Next, in few-shot training (Phase 2), $f_\theta$ is duplicated to construct a Siamese network, and the model is further trained using triplet loss to refine the embedding space. Finally, in the fine-tuning phase (Phase 3), inspired by Chen \textit{et al}. \cite{meta-baseline}, the network is fine-tuned with a few examples of novel exercise data obtained during the exercise registration procedure (\ref{sec:3-a}). This fine-tuning improves the model’s ability to adapt to new exercises.

In the test step, $f_\theta$ is used to count repetitions for a new exercise through the following process: The user first registers 5 repetitions by following the system's instructions (\ref{sec:3-a}). Based on the registered signals and annotations, windows at the peak points are selected as positive samples, and windows at the non-peak points are selected as negative samples. As the user performs the exercise, real-time data is collected and treated as anchor samples, which are then compared with the positive and negative samples to classify peak and non-peak points. Finally, the transition counting function determines the total number of repetitions, and the system displays the results in real-time.

\section{Experiments}
\label{sec:4}
In this section, we provide a comprehensive overview of our experimental setup. Specifically, we explain the structure of the dataset, the network architecture, and the loss function used in each phase. Finally, we discuss the post-processing method employed to determine the repetitions.

\begin{figure*}[t]
    \centering
    \includegraphics[width=0.9\linewidth]{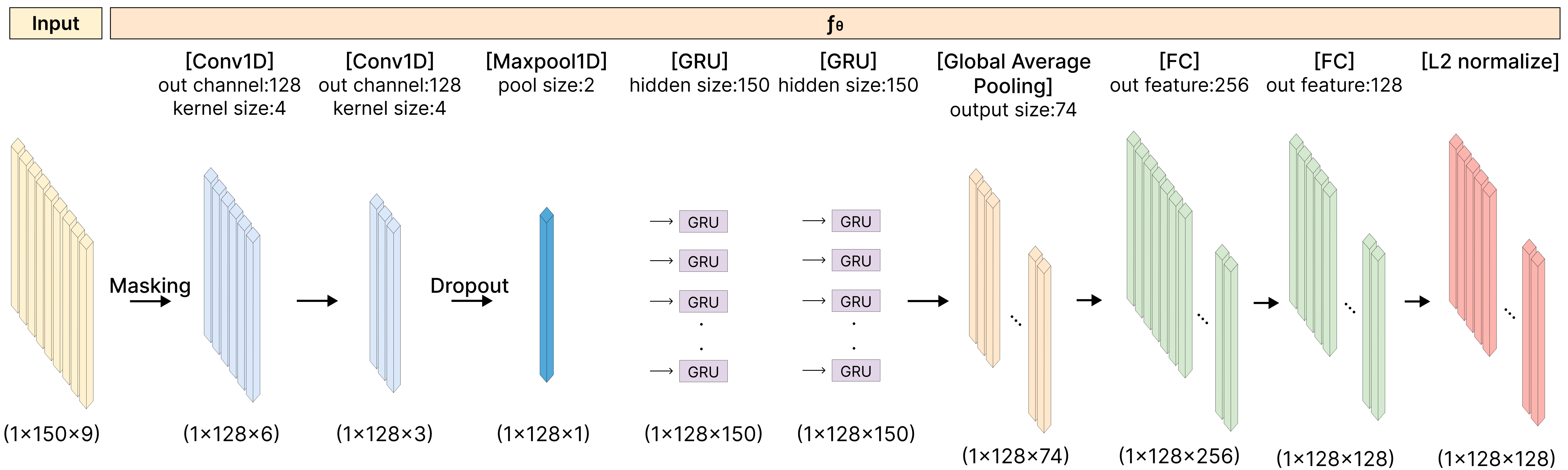}
    \caption{Architecture of $f_\theta$: Masking, 1D Conv, Dropout, Max pooling, GRU, GAP, FC, and L2 normalization}
    \label{fig: f_theta}
\end{figure*}

\subsection{Dataset}
The dataset used in this study is comprised of IMU sensors and audio sound data generated during exercises and was collected from 11 healthy participants under the IRB approval by IRB SSU-202202-HR-414-1 at Soonsil University. In the data collection process, a single IMU sensor and voice recorder using a smart earbud are attached to the ear of a participant. Participants were asked to perform 30 different exercises repeatedly for each, and sensor and audio data were collected. A detailed description of the dataset and the process of data collection experiments are provided in our previous research \cite{har_sensor_fusion}.

For this study, we used only the IMU data, which includes accelerometer, gyroscope, and magnetometer readings in the x, y, and z directions at a 92 Hz rate. Among the 30 exercises conducted in the previous work, we selected 28 types of weight training and home workouts due to the lack of clear sensor patterns in the other two exercises. Each exercise is labeled with two labels, peak, non-peak. The total number of movements was 19,777, with an average of 706 repetitions for each exercise performed in the experiment. Table \ref{tab:parameter} presents the workouts and their statistics.

Our dataset exhibits significant intra-variation within the peak and non-peak labels. For example, the peak signal of the Burpee exercise quite differs considerably from that of a Squat, reflecting the inherent variability in movement patterns across different exercises.

\subsection{Preprocessing}
The IMU sensor data were preprocessed and segmented using the sliding window method. Each segment (window) $x_i = (a_i, g_i,m_i)$ consists of 3-axis accelerations $a_i\in R^{T \times3}$, 3-axis angular rates $g_i\in R^{T \times3}$, and 3-axis magnetic fields $m_i\in R^{T \times3}$, where $T$ denotes the window size. 
Table \ref{tab:parameter} shows the optimal sliding window size and stride for each exercise, based on our previous work, which developed separate models for individual exercises.

\begin{figure}[h]
    \centering
    \includegraphics[width=1.0\linewidth]{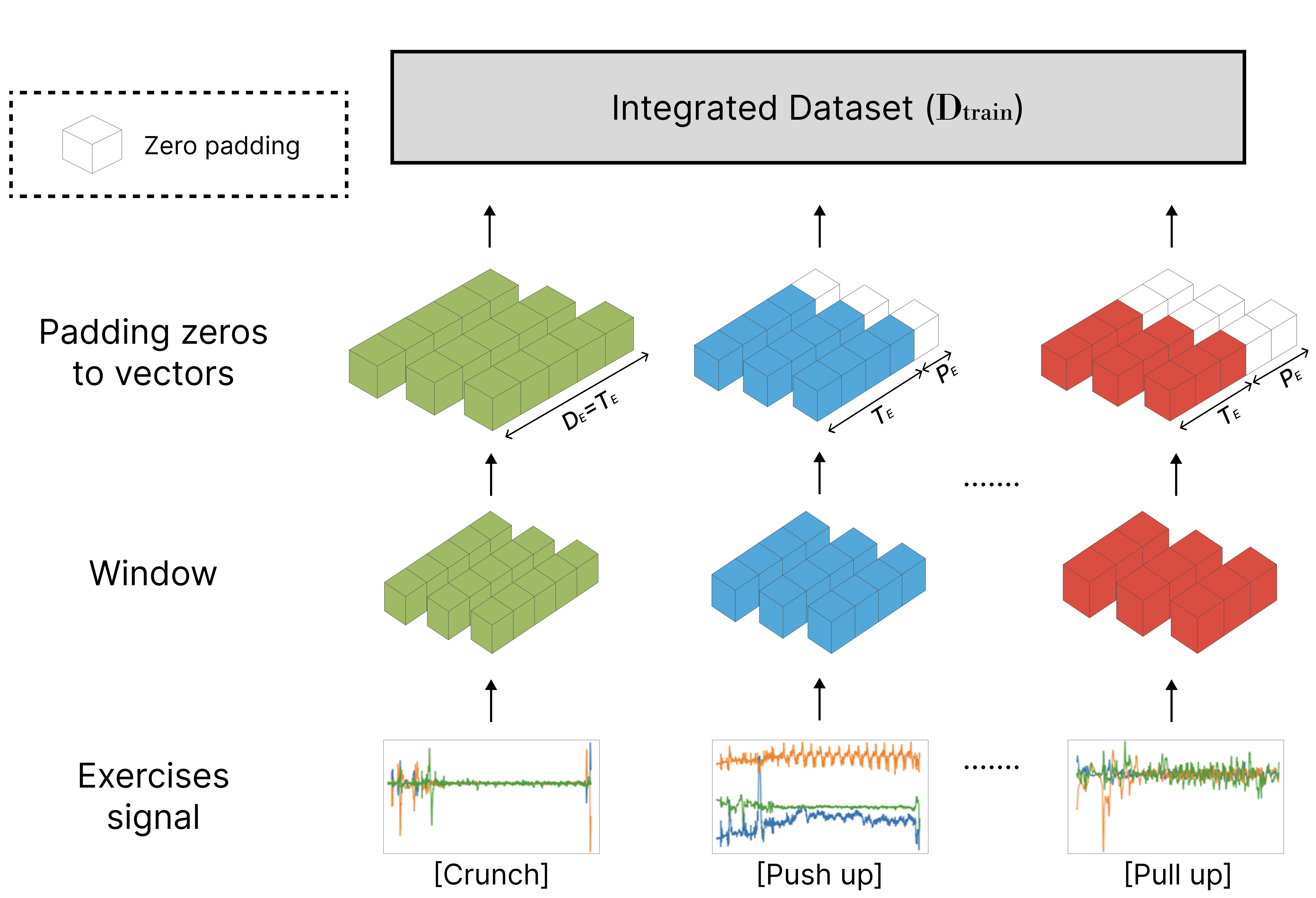}
    \caption{Padding process}
    \label{fig: padding}
\end{figure}

To create a unified dataset ($D_{train}$) that integrates all exercise types, we addressed the issue of varying window sizes and strides. When combining data with different window lengths, some windows were too short for the model’s input requirements. To resolve this, we applied zero padding to windows with shorter lengths, making all windows consistent with the longest window size. Fig. \ref{fig: padding} illustrates this process, where $D_E$ represents the input window size, $T_E$ is the optimal window size for each exercise type, and $P_E$ indicates the padding added.

As a result, we constructed a comprehensive dataset that effectively captures the distinct characteristics of each workout. This unified dataset enables efficient training and supports the development of a robust model capable of accurately classifying and counting repetitions across a wide range of exercises.

\begin{table}[t]
\caption{Statistics of workouts and best window size with stride}
\resizebox{\columnwidth}{!}{%
\renewcommand{\arraystretch}{1.3}
{\small
\begin{tabular}{c|c|cccc}
\hline
\textbf{ID} & \textbf{Exercise} & \textbf{Reps} & \textbf{\begin{tabular}[c]{@{}c@{}}Time per \\ Reps (sec)\end{tabular}} & \textbf{\begin{tabular}[c]{@{}c@{}}Window \\ size\end{tabular}} & \textbf{Stride} \\ \hline \hline
1 & Squat & 809 & 2.94 & 80 & 20 \\
2 & Dips & 663 & 1.48 & 50 & 25 \\
3 & Pull up & 812 & 2.2 & 80 & 10 \\
4 & Fixed Lunge & 1120 & 2.16 & 80 & 50 \\
5 & Barbell Row & 875 & 1.42 & 80 & 10 \\
6 & Push Up & 489 & 0.89 & 50 & 25 \\
7 & Overhead Press & 1065 & 2.38 & 100 & 10 \\
8 & Deadlift & 785 & 2.89 & 150 & 10 \\
9 & Crunch & 670 & 0.9 & 50 & 10 \\
10 & Back Extension & 718 & 2.18 & 100 & 20 \\
11 & Thruster & 529 & 2.92 & 50 & 25 \\
12 & Calf Raises & 844 & 1.59 & 80 & 10 \\
13 & Lat Pull Down & 634 & 2.22 & 50 & 10 \\
14 & Barbell Curl & 673 & 1.92 & 100 & 10 \\
15 & Side Lateral Raise & 685 & 1.67 & 100 & 20 \\
16 & Arm Pull Down & 737 & 2.33 & 80 & 20 \\
17 & Seated Row & 639 & 2.01 & 80 & 20 \\
18 & Cable Push Down & 686 & 1.7 & 100 & 10 \\
19 & Mountain Climber & 1149 & 0.76 & 50 & 10 \\
20 & Side Lunge & 490 & 3.63 & 80 & 10 \\
21 & Step Lunge & 669 & 4.34 & 150 & 10 \\
22 & Standing Cross-Knee Up & 463 & 2.4 & 50 & 10 \\
23 & Crab Toe Touch & 529 & 2.54 & 50 & 10 \\
24 & Jumping Squat & 933 & 1.42 & 50 & 10 \\
25 & Arm Walking Push Up & 431 & 6.11 & 150 & 10 \\
26 & Jumping Jack & 658 & 2.04 & 150 & 10 \\
27 & Burpee & 680 & 2.52 & 80 & 10 \\
28 & Lunge with Rotation & 342 & 5.58 & 100 & 10 \\\hline
\end{tabular}%
}}
\label{tab:parameter}
\end{table}

\subsection{Few-shot learning for repetition counting}

\textbf{Phase 1: Standard Classification Training} The first phase is the classification training, which aims to obtain an initial weight for the few-shot training phase. For this, we utilize $D_{train}$ for training, which consists of peak and non-peak data integrated from 27 different exercises.

The structure of $f_\theta$ is a hybrid neural network that combines 1D CNN and GRU, as shown in Fig. \ref{fig: f_theta}, making it suitable for time-series data. The network $f_\theta$ consists of masking, 1D convolutional layers, dropout, max pooling, GRU, global average pooling (GAP), fully connected (FC) layers, and L2-normalization. As depicted in Fig. \ref{fig: padding}, padding is applied to variable-sized windows to create a uniform dataset. Since the padding contains no actual information, we use a masking layer to remove the zero padding from computations. The output tensor $\Tilde{x}_i$ of the $f_\theta$ network is computed as follows:

\begin{equation}
\label{eq: phase1_prediction}
    \Tilde{x}_i = f_{\theta}(x_i)
\end{equation}

For binary classification, a linear layer with a sigmoid activation function is added to $f_\theta$, and binary cross-entropy is used as the loss function to predict peak or non-peak points:

\begin{equation}
\label{eq: phase1_loss}
\begin{array}{cl}
    \hat{y}_i = Sigmoid(W_o\Tilde{x}_i + b_o) \\
    L_{phase1}=-\frac{1}{N}\sum\limits_{i=1}^N y_i log(\hat{y}_i)+(1-y_i)log(1-\hat{y}_i)
\end{array}
\end{equation}

\textbf{Phase 2: Few-Shot Training} The second phase involves the few-shot training using a Siamese network. The objective is for $f_\theta$ to learn data representations that can generalize to new, unseen exercises. The Siamese network is built by duplicating $f_\theta$, excluding the linear layer added for classification in Phase 1. The weights learned in Phase 1 are used as the initial parameters, and the network is trained without freezing any layers.

For this phase, we transform the input into triplet pairs. When the peak point of a behavior signal serves as the anchor ($x^a$), the other data representing the same peak is the positive ($x^p$), and the data representing the non-peak point is the negative ($x^n$). Conversely, when the anchor is a non-peak point, the positive sample is another non-peak point, and the negative is a peak point.

To avoid inefficiency in using all possible triplet combinations, we apply semi-hard triplet mining \cite{facenet}. This method selects the negative sample ($x^n$) such that it is not closer to the anchor ($x^a$) than the positive sample ($x^p$), and the distance between $x^a$ and $x^n$ does not exceed the distance between $x^a$ and $x^p$ by more than a margin ($\alpha$), as shown in Eq. (\ref{eq: semihard_triplet}). This ensures that extremely easy or difficult samples are excluded, focusing the training on moderately challenging examples. The margin $\alpha$ is set to 1.0.

\begin{equation}
\label{eq: semihard_triplet}
\begin{array}{cl}
    {\Vert f_\theta(x^a)-f_\theta(x^p) \Vert}^2<{\Vert f_\theta(x^a)-f_\theta(x^n) \Vert}^2\\ <{\Vert f_\theta(x^a)-f_\theta(x^p) \Vert}^2 + \alpha
\end{array}
\end{equation}

The triplet loss, defined in Eq. (\ref{eq: triplet_loss}), aims to bring the positive pair closer in the embedding space and push the negative pair farther apart. The hinge loss function $[\cdot]_+$ ensures that only the violations of the margin contribute to the loss, meaning that when the negative sample is already sufficiently far from the anchor, no additional penalty is applied.

\begin{equation}
\label{eq: triplet_loss}
    L_{triplet}(t)=[{\Vert f_\theta(x^a)-f_\theta(x^p) \Vert}^2 - {\Vert f_\theta(x^a)-f_\theta(x^n) \Vert}^2 + \alpha]_+
\end{equation}

The total triplet loss across all triplet pairs is averaged, as shown in Eq. (\ref{eq:triplet_loss_2}), to optimize the model’s performance on the dataset.

\begin{equation}
\label{eq:triplet_loss_2}
    L_{phase2} = \frac{1}{N} \sum\limits_{i=1}^N L_{triplet}(t_i)
\end{equation}

\textbf{Phase 3: Few-Shot Fine-Tuning} The third phase involves few-shot fine-tuning to adapt $f_\theta$ to novel exercise movements. We use the same network and loss function as in Phase 2. However, unlike in Phase 2, we freeze all layers in the feature extractor except for the last two fully-connected layers, which are the only ones trained in this phase to adjust to the novel exercises.

To perform few-shot classification for novel movements, we first obtain a triplet pair for the new exercise signal. Through the registration procedure (\ref{sec:3-a}), we acquire novel exercise signals consisting of five repetitions of data ($D_{register}$).

Prior knowledge about the window size and stride in $D_{register}$ is needed to apply the sliding window technique. To address this, we propose a parameter setting rule based on exercise duration. As shown in Table \ref{tab:parameter}, exercises with shorter durations use smaller window sizes, while longer exercises use larger windows. Specifically, we set a window size of 100 and stride of 50 for exercises lasting longer than 1.5 seconds, and a window size of 50 with a stride of 25 for shorter exercises.

The sliding window technique provides an effect similar to data augmentation without the need for additional augmentation methods. By applying this technique to $D_{register}$, we naturally obtain more than five windows containing both peak and non-peak points. These augmented 5-shot signals are then used for fine-tuning.

In this phase, we use a small number of epochs (15) and a learning rate of 5e-5 to prevent overfitting, which is critical for improving performance.

\textbf{Post-processing} In this section, we discuss how classification results are determined and repetitions are counted. Using the $f_\theta$ obtained from Phase 3, we compare the similarities between two pairs, $(x^a, x^p)$ and $(x^a, x^n)$, and classify $x^a$ into the class with the higher similarity. The cosine similarity of the positive pair ($SP$) and the negative pair ($SN$) is calculated using Eq. (\ref{eq:prediction}) to determine peak and non-peak points. $x^p$ and $x^n$ are randomly selected from the set constructed from $D_{register}$, and the final similarity is averaged over five samples. The class is then determined by comparing $SP$ and $SN$.

\begin{equation}
\label{eq:prediction}
\begin{array}{cl}
    SP=\frac{x^a \cdot x^p}{\Vert x^a \Vert \Vert x^p \Vert} \quad SN=\frac{x^a \cdot x^n}{\Vert x^a \Vert \Vert x^n \Vert} \\
    \\
    \hat{y}= \begin{cases} \textrm{peak},\;\textrm{if}\;SP\geq SN\\
    \textrm{non-peak},\;\textrm{if}\;SP < SN
    \end{cases}

\end{array}
\end{equation}

The network outputs a series of peak and non-peak points, where peaks indicate the start of a repetition. We apply a transition counting function (\ref{sec:3-d}) and count the total number of repetitions. 

\subsection{Implementation details}
We utilized the same network structure across all phases of the experiment. For training, data from 27 workouts were used, while one workout was reserved for testing. This testing process was repeated for each of the 28 workouts to ensure thorough evaluation. The window size and stride for novel exercises were determined using the parameter setting rule outlined in Phase 3, maintaining consistency in data processing throughout the experiments.

For evaluation, we employed standard classification metrics such as accuracy, recall, precision, and F1 score to comprehensively assess the model’s performance. During the training phase, we used the Adam optimizer with a learning rate of 1e-3 and a weight decay of 1e-4, which ensured effective learning while minimizing overfitting. In the fine-tuning phase, the Rectified Adam optimizer was applied with a learning rate of 5e-5, specifically targeting the fully connected layers to allow precise adjustments when adapting to novel exercise movements.

\section{Results}
\label{sec:5}

\begin{table}[t]
\caption{Test performance for each exercise}
\resizebox{\columnwidth}{!}{%
\renewcommand{\arraystretch}{1.6}
{\small
\begin{tabular}{c|c|cccc}
\toprule[1.5pt] 
\textbf{ID} &\textbf{Exercise} &\textbf{Accuracy} &\textbf{Recall} &\textbf{Precision} &\textbf{F1}\\ \hline\hline
1  & Squat                  & 0.90     & 0.77   & 0.87      & 0.81 \\ 
2  & Dip                    & 0.93     & 0.90   & 0.84      & 0.88 \\ 
3  & Pull-Up                & 0.84     & 0.79   & 0.82      & 0.81 \\ 
4  & Fixed Lunge            & 0.88     & 0.78   & 0.92      & 0.85 \\ 
5  & Barbell Row            & 0.80     & 0.67   & 0.56      & 0.61 \\ 
6  & Push-Up                & 0.82     & 0.96   & 0.68      & 0.80 \\ 
7  & Overhead Press         & 0.83     & 0.49   & 0.75      & 0.59 \\ 
8  & Deadlift               & 0.64     & 0.84   & 0.40      & 0.55 \\ 
9  & Crunch                 & 0.53     & 0.95   & 0.45      & 0.61 \\ 
10 & Back Extension         & 0.70     & 0.99   & 0.52      & 0.68 \\ 
11 & Thrust                 & 0.91     & 0.91   & 0.86      & 0.89 \\ 
12 & Calf Raise             & 0.54     & 1.00   & 0.54      & 0.70 \\
13 & Lat Pull-Down          & 0.73     & 0.70   & 0.57      & 0.63 \\ 
14 & Barbell Curl           & 0.67     & 0.87   & 0.53      & 0.66 \\ 
15 & Side Lateral Raise     & 0.86     & 0.76   & 0.96      & 0.85 \\ 
16 & Arm Pull-Down          & 0.83     & 0.81   & 0.68      & 0.74 \\
17 & Seated Row             & 0.70     & 0.81   & 0.59      & 0.69 \\ 
18 & Cable Push-Down        & 0.63     & 0.73   & 0.56      & 0.63 \\ 
19 & Mountain Climbers      & 0.64     & 0.44   & 0.65      & 0.53 \\ 
20 & Side Lunge             & 0.84     & 0.80   & 0.88      & 0.84 \\ 
21 & Step Lunge             & 0.90     & 0.82   & 0.89      & 0.85 \\ 
22 & Standing Cross Knee Up & 0.93     & 0.87   & 0.85      & 0.86 \\ 
23 & Crap Toe Touch         & 0.88     & 0.86   & 0.80      & 0.83 \\ 
24 & Jumping Squat          & 0.81     & 0.70   & 0.93      & 0.80 \\ 
25 & Arm Walk Push-Up       & 0.74     & 0.54   & 0.71      & 0.62 \\ 
26 & Jumping Jack           & 0.82     & 0.78   & 0.91      & 0.84 \\ 
27 & Burpee                 & 0.90     & 0.94   & 0.92      & 0.93 \\ 
28 & Lunge Rotation         & 0.68     & 0.48   & 0.74      & 0.59 \\ \bottomrule[1.5pt]
\end{tabular}%
}}
\label{tab:result_classification}
\end{table}

\subsection{Results on 5-shot classification}
Table \ref{tab:result_classification} presents the 5-shot classification test results. All data signals from the test workouts were used for testing, and peak classification was performed on the novel workouts using the fine-tuned model.

F1 scores for exercises such as squats, dips, pull-ups, fixed lunges, push-ups, thrusters, side lateral raises, side lunges, step lunges, standing cross-knee ups, crab toe touches, jumping squats, jumping jacks, and burpees were relatively high, scoring 0.80 or above. In contrast, exercises like barbell rows, overhead presses, deadlifts, crunches, back extensions, lat pull-downs, barbell curls, cable push-downs, mountain climbers, arm walk push-ups, and lunge rotations had lower F1 scores, typically 0.70 or below.

Table \ref{tab:result_stride} illustrates the differences in test performance when varying the window size and stride for barbell rows and squats. 
For squats, a window size of 100 and stride of 50 were set according to the established rule for the test, while the preliminary parameters (window size: 80, stride: 20) were provided in Table \ref{tab:parameter}. When using the preliminary information, the F1 value for barbell rows was 0.1 higher, whereas for squats, it was only 0.02 higher. The overall scores that involve accuracy, recall, and precision increased.

\begin{table}[t]
\centering
\caption{The test performance, according to the change in window size and stride}
\resizebox{\columnwidth}{!}{%
\renewcommand{\arraystretch}{1.6}
{\small
\begin{tabular}{c|cc|cc}
\toprule[1.5pt]
Exercise &\multicolumn{2}{c|}{Barbell row} &\multicolumn{2}{c}{Squat} \\ \cline{1-5} 
Window Size / Stride    & 100/50      & 80/10       & 100/50       & 80/20  \\ \hline\hline
Accuracy  & 0.80            & 0.85           & 0.90         & 0.90       \\
Recall    & 0.67            & 0.73           & 0.77         & 0.76       \\
Precision & 0.56            & 0.70           & 0.87         & 0.89       \\
F1        & 0.61            & 0.71           & 0.81         & 0.83       \\ 
\bottomrule[1.5pt]
\end{tabular}%
}}
\label{tab:result_stride}
\end{table}

\begin{table}[h!]
\centering
\caption{Error ratio in the repetition sets}
\resizebox{\columnwidth}{!}{%
\renewcommand{\arraystretch}{1.6}
\begin{tabular}{c|c|c|ccccccc}
\toprule[1.5pt]
 & & & \multicolumn{7}{c}{Error ratio(\%)}\\ \cline{4-10} 
ID&Workout & \begin{tabular}[c]{@{}c@{}}Total\\ sets\end{tabular} & e$|0|$ & e$|1|$ & e$|2|$ & e$|3|$ & e$|4|$ & e$|5|$ & e$|\!>\!5|$ \\ \hline\hline
1&Squat       & 79 & 12.7 & 3.8 & 29.1 & 13.9 & 16.5 & 11.4 & 12.7 \\
2&Dip         & 52 & 30.8 & 15.4 & 7.7 & 7.7 & 15.4 & 7.7 & 15.4 \\
3&Pull-Up     & 85 & 15.3 & 20.0 & 15.3 & 17.6 & 2.4 & 16.5 & 12.9 \\
4&Fixed Lunge & 78 & 10.3 & 15.4 & 26.9 & 19.2 & 6.4 & 5.1 & 16.7 \\
5&Barbell Row & 60 & 3.3 & 1.7 & 16.7 & 15.0 & 13.3 & 33.3 & 16.7 \\
6&Push-Up     & 17 & 17.6 & 41.2 & 5.9 & 0 & 35.3 & 0 & 0 \\
7&Overhead Press & 85 & 4.7 & 3.5 & 14.1 & 5.9 & 15.3 & 35.3 & 21.2 \\
8&Deadlift    & 60 & 0 & 18.3 & 13.3 & 8.3 & 20.0 & 23.3 & 16.7 \\
9&Crunch      & 21 & 14.3 & 4.8 & 14.3 & 0 & 23.8 & 33.3 & 9.5 \\
10&Back Extension & 48 & 10.4 & 4.2 & 6.3 & 6.3 & 20.8 & 33.3 & 9.5 \\
11&Thrust      & 36 & 22.2 & 33.3 & 2.8 & 11.1 & 16.7 & 8.3 & 5.6 \\
12&Calf Raise  & 47 & 0 & 19.1 & 14.9 & 23.4 & 17 & 14.9 & 10.6 \\
13&Lat Pull-Down & 46 & 13.0 & 4.3 & 8.7 & 19.6 & 23.9 & 21.7 & 8.7 \\
14&Barbell Curl & 53 & 9.4 & 18.9 & 22.6 & 18.9 & 7.5 & 5.7 & 17.0 \\
15&Side Lateral Raise & 47 & 27.7 & 10.6 & 12.8 & 2.1 & 21.3 & 17.0 & 8.5 \\
16&Arm Pull-Down & 54 & 7.4 & 13.0 & 24.1 & 9.3 & 3.7 & 5.6 & 37.0 \\
17&Seated Row  & 47 & 17.0 & 21.3 & 8.5 & 10.6 & 2.1 & 31.9 & 8.5 \\
18&Cable Push-Down & 47 & 6.4 & 0 & 31.9 & 6.4 & 6.4 & 40.4 & 8.5 \\
19&Mountain Climbers & 34 & 0 & 8.8 & 23.5 & 8.8 & 29.4 & 20.6 & 8.8 \\
20&Side Lunge  & 32 & 31.3 & 25.0 & 12.5 & 9.4 & 9.4 & 6.3 & 6.3 \\
21&Step Lunge  & 55 & 30.9 & 9.1 & 21.8 & 9.1 & 12.7 & 9.1 & 7.3 \\
22&Standing Cross Knee Up & 33 & 6.1 & 9.1 & 15.2 & 48.5 & 6.1 & 3.0 & 12.1 \\
23&Crab Toe Touch & 31 & 22.6 & 9.7 & 22.6 & 9.7 & 9.7 & 19.4 & 6.5 \\
24&Jumping Squat & 37 & 24.3 & 13.5 & 27.0 & 10.8 & 8.1 & 5.4 & 10.8 \\
25&Arm Walk Push-Up & 26 & 3.8 & 11.5 & 15.4 & 11.5 & 23.1 & 23.1 & 11.5 \\
26&Jumping Jack & 39 & 10.3 & 41 & 0 & 25.6 & 7.7 & 5.1 & 10.3 \\
27&Burpee      & 41 & 19.5 & 17.1 & 31.7 & 4.9 & 12.2 & 9.8 & 4.9 \\
28&Lunge Rotation & 20 & 5.0 & 0 & 10.0 & 5.0 & 25.0 & 30.0 & 25.0 \\
\bottomrule[1.5pt]
\end{tabular}%
}
\label{tab:result_counting}
\end{table}

\subsection{Results on repetition counting}
Table \ref{tab:result_counting} presents the results of repetition counting for individual exercises in terms of the number of errors compared to the actual count using the Siamese network. The columns indicate the repetition error ratio. ``e$|X|$" denotes the number of exercise sets with the ``$|X|$'' repetition error count. ``$|X|$'' represents the absolute error count of 0, 1, 2, 3, 4, 5, or more than five errors. In our dataset, the mean repetitions per one set is fifteen. The percentage of errors more than five times is 13.2\%, and the rate of error-free counting is 12.9\%. The probability of correctly calculating ten or more repetitions in one set is 86.8\%. Generally, nine repetitions in one set were counted accurately.

\begin{figure}[t]
    \centering
    \includegraphics[width=1.0\linewidth]{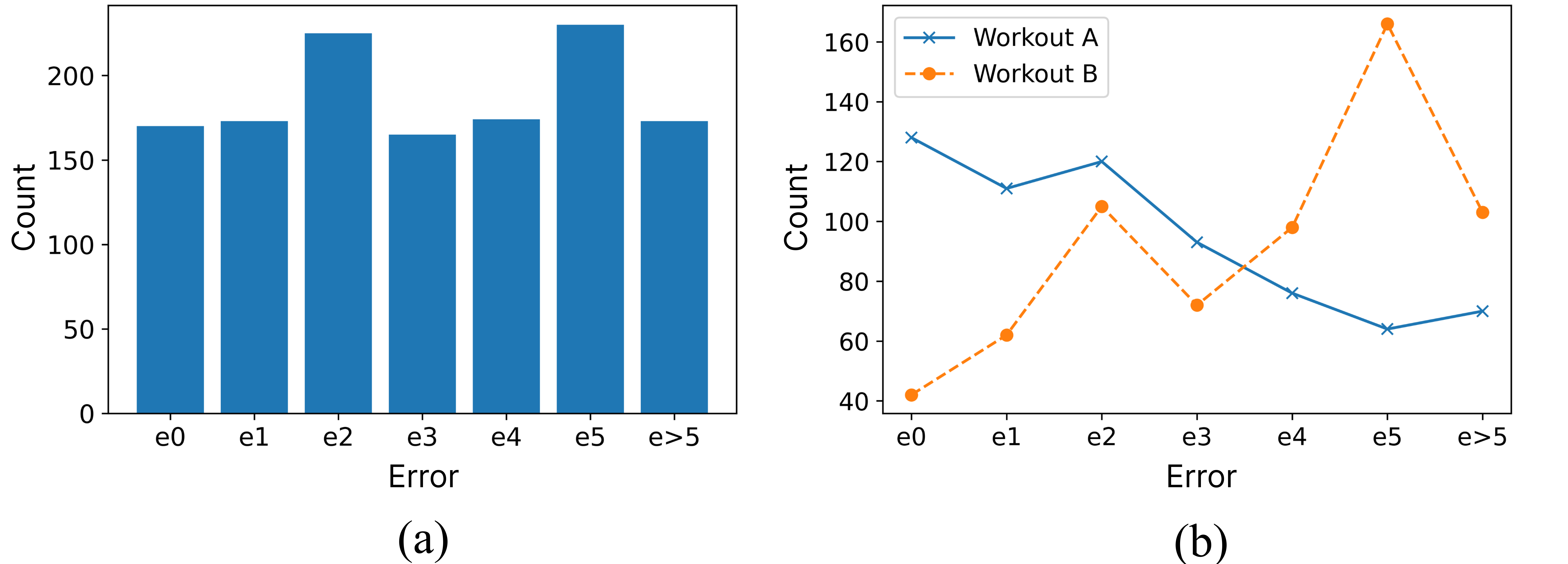}
    \caption{(a) total subjects by error; (b) error subjects by workout performance; Workout A: high-performance; Workout B: low-performance in Table \ref{tab:result_classification}}
    \label{fig:result_repetition}
\end{figure}

Workouts with high classification performance showed a significant number of subjects corresponding to error counts of $e|0|$, $e|1|$, and $e|2|$. In contrast, workouts with lower performance had more subjects corresponding to higher error counts of $e|3|$, $e|4|$, and $e|5|$ (Fig. \ref{fig:result_repetition}). However, low classification performance did not necessarily imply that all subjects had high incorrect counting rates. In cases where performance was low, such as mountain climber and deadlift exercises, subjects corresponding to $e|1|$ error counts were often present. Even if the performance of classifying peak frames was slightly low, high performance in counting could be achieved due to the transition counting function, as long as the classification of non-peak frames was accurate and the peak frames were appropriately classified. As a result, recall played a more critical role in accurate repetition counting.

\section{Discussion}
\label{sec:6}
\begin{figure*}[t]
    \centering
    \includegraphics[width=1.0\linewidth]{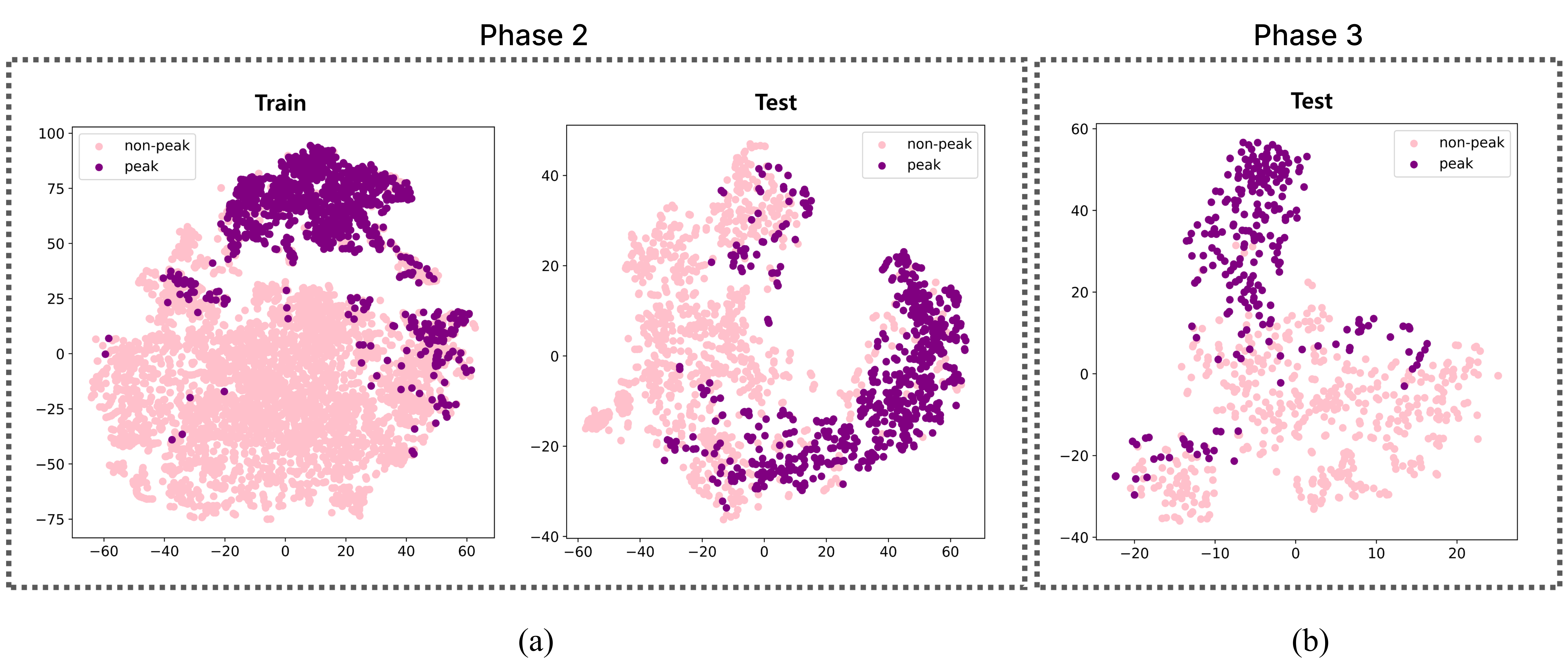}
    \caption{Embedding space distribution for train and test data; side lunge as test exercise; (a) Phase 2: few-shot training; (b) Phase 3: few-shot fine-tuning}
    \label{fig:result_tsne}
\end{figure*}

In this paper, we propose a repetition counting framework for few-shot tasks using a Siamese network that employs triplet loss and fine-tuning techniques in situations where only one IMU sensor is used. Our classification results demonstrate that the proposed framework performs effectively in few-shot tasks. The embedded distribution results, which are well-classified according to the class as shown in Fig. \ref{fig:result_tsne}, indicate that the Siamese network possesses high representational power. The fine-tuning phase is also effective in refining the embedding space. Table \ref{tab:result_stride} highlights the importance of the hyperparameter rule. Exercises with relatively low performance, such as the barbell rows, exhibit greater sensitivity to hyperparameter settings, showing the close relationship between improved performance and optimal hyperparameter tuning.

\subsection{Baseline for few-shot repetition counting}
Several studies have implemented deep learning for repetition counting using IMU sensors \cite{har_rc_exercise, har_rc_rehabilitation, har_rc_workout}. However, no study has developed a deep learning-based model for novel exercises. Our objective is to accurately count repetitions for new exercises.

Workouts with precise head movements or a substantial gap between repetitions exhibit relatively high performance. In contrast, workouts with narrow intervals between repetitions and rapid actions yield low performance (see Table \ref{tab:result_classification}). These results may be attributed to the fact that most training set workouts have a relatively long repetition duration and a clear gap between repetitions. Specifically, the arm walking push-up, which has a considerably long repetition time, displayed significantly different characteristics compared to the training set, resulting in lower performance despite ample head movement and a wide gap between repetitions.

Previous studies have developed individual models for each workout to count repetitions. While this approach exhibits high performance in separate exercises, the number of weight training events is immense, exceeding 90. Building individual models for all movements is challenging in this context. From this perspective, our framework, which determines the optimal window for each exercise, integrates the dataset for each activity, and addresses problems through few-shot learning, may serve as a baseline for future few-shot repetition counting studies.

\subsection{The advantage of few-shot learning}
Peak data refers to the maximum point in the exercise cycle, such as the top of a squat or the highest point of a push-up, and appears semantically straightforward. However, the peak patterns of each activity can vary significantly, leading to challenges in predicting novel data. Traditional classification models that divide peak and non-peak points have limitations when faced with classes that exhibit large intra-variations. In contrast, the metric-based few-shot learning method proposed in this study classifies based on the distance between the embedded features of the frames, offering a flexible approach that adjusts the decision boundary dynamically according to the variations in peak patterns.

\section{Conclusion}
\label{sec:7}
This study presents a pioneering few-shot repetition-counting method tailored for sensor-based exercise tracking, marking a significant step forward in adapting to new, unseen exercises. By employing pre-trained initial weights and fine-tuning techniques, our framework demonstrates the potential for effective adaptation to novel exercises using only a single IMU sensor.

Our research has shown that combining metric-based learning with appropriate hyperparameter settings can significantly improve performance when classifying exercises with large intra-variations. Furthermore, our framework, which integrates data across various exercises and leverages few-shot learning, establishes a solid foundation for future few-shot repetition counting studies.

In future work, we aim to explore various few-shot repetition counting methods, including applying different techniques such as ProtoNet \cite{proto}, MAML \cite{maml}, and Meta-Opt \cite{metaopt}, which have shown effective results in other domains. Additionally, we plan to develop more robust and adaptive methods that can accurately count repetitions across an even broader range of exercises.

\printcredits

\bibliographystyle{cas-model2-names}

\bibliography{manuscript}

\end{document}